\documentclass[conference]{IEEEtran}
\IEEEoverridecommandlockouts
\usepackage{cite}
\usepackage{amsmath,amssymb,amsfonts}
\usepackage{algorithmic}
\usepackage{graphicx}
\usepackage{textcomp}
\usepackage{balance}
\usepackage{booktabs} 
\usepackage{bm}
\usepackage{array}
\usepackage{url}
\usepackage[ruled,linesnumbered]{algorithm2e}
\usepackage{xcolor}
\def\BibTeX{{\rm B\kern-.05em{\sc i\kern-.025em b}\kern-.08em
    T\kern-.1667em\lower.7ex\hbox{E}\kern-.125emX}}
\begin{document}

\title{Alignahead: Online Cross-Layer Knowledge Extraction on Graph Neural Networks\\
}

\author{\IEEEauthorblockN{Jiongyu Guo, Defang Chen, Can Wang\thanks{
Can Wang is the Corresponding author.}}
\IEEEauthorblockA{
\textit{Zhejiang University},
\textit{ZJU-Bangsun Joint Research Center}\\
\textit{Shanghai Institute for Advanced Study of Zhejiang University}\\
%Hangzhou, China \\
\{jy.guo, defchern, wcan\}@zju.edu.cn}
}

\maketitle

\begin{abstract}
Existing knowledge distillation methods on graph neural networks (GNNs) are almost offline, where the student model extracts knowledge from a powerful teacher model to improve its performance. However, a pre-trained teacher model is not always accessible due to training cost, privacy, etc. 
In this paper, we propose a novel online knowledge distillation framework to resolve this problem. Specifically, each student GNN model learns the extracted local structure from another simultaneously trained counterpart in an alternating training procedure. We further develop a cross-layer distillation strategy by \textit{aligning ahead} one student layer with the layer in different depth of another student model, which theoretically makes the structure information spread over all layers. 
Experimental results on five datasets including PPI, Coauthor-CS/Physics and Amazon-Computer/Photo demonstrate that the student performance is consistently boosted in our collaborative training framework without the supervision of a pre-trained teacher model.
In addition, we also find that our \textit{alignahead} technique can accelerate the model convergence speed and its effectiveness can be generally improved by increasing the student numbers in training. Code is available: \url{https://github.com/GuoJY-eatsTG/Alignahead}
\end{abstract}


\section{Introduction}
Deep neural networks (DNNs) have achieved great success in various computer vision \cite{He2016DeepRL}, natural language processing \cite{Devlin2019BERTPO} and speech recognition \cite{Dahl2012ContextDependentPD} tasks. However, those well-performed DNNs typically require very high computation and large memory usage, making them difficult to be deployed on platforms with limited resources. 
To solve this problem, knowledge distillation was proposed as one of the mainstream model compression methods \cite{bucilua2006model,ba2014deep,hinton2015distilling,romero2015fitnets,zagoruyko2017paying,chen2021cross,zhang2021confidence,chen2022simkd}. Knowledge distillation (KD) aims to transfer knowledge from a pre-trained, cumbersome and powerful teacher model to a compressed student model with fewer parameters, hoping that the student model can achieve better performance than training with labels alone and even be on par with the teacher model.
\begin{figure}[ht]
  \centering
  \includegraphics[width=\linewidth]{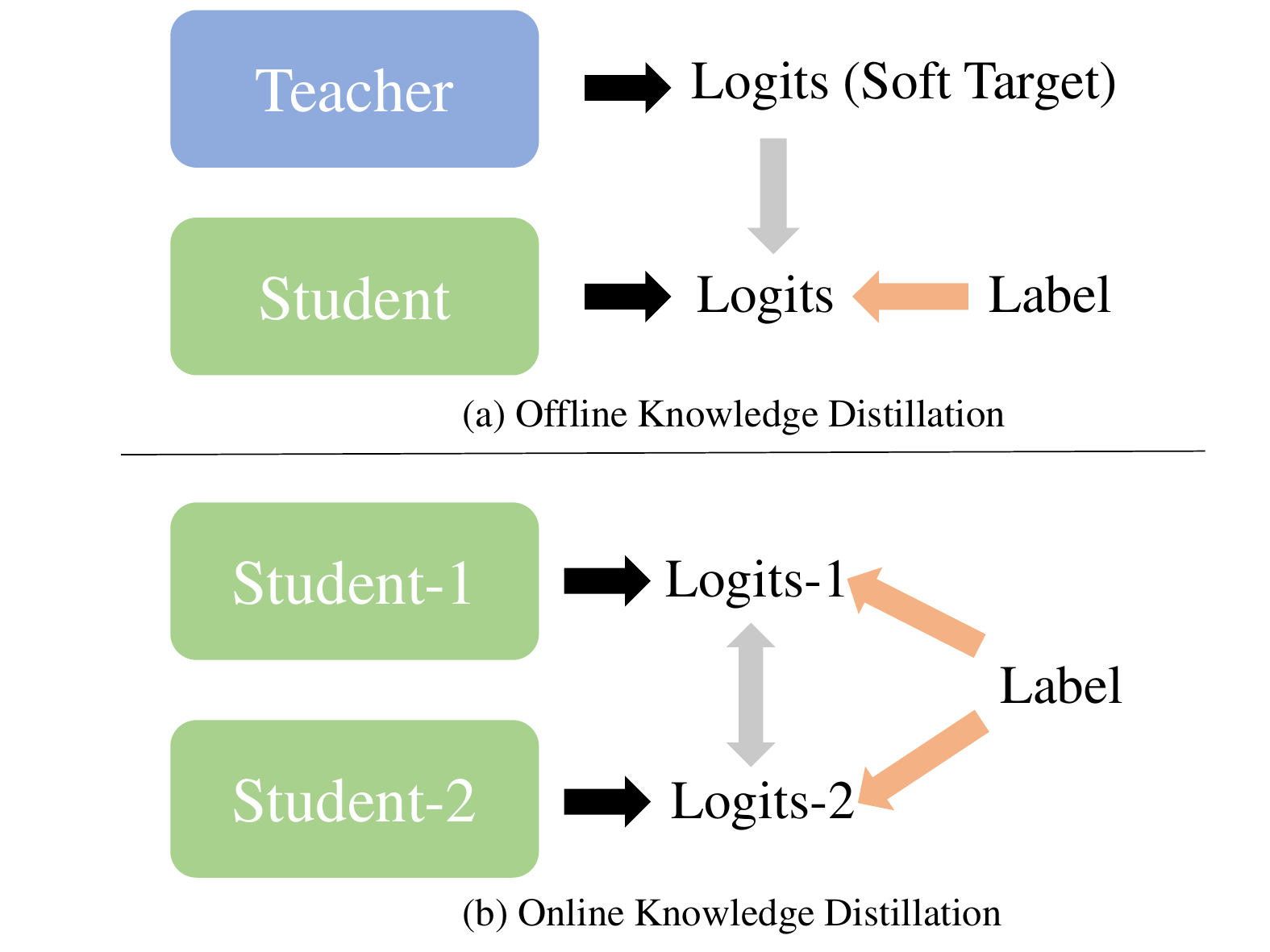}
  \caption{(a) Offline KD: knowledge is transferred from the pre-trained and fixed teacher model to the student model. (b) Online KD: two student models are trained from scratch by learning from each other alternately.}
  \label{fig:overview}
\end{figure}

However, the existing KD methods mainly focus on convolutional neural networks rather than graph neural networks (GNNs). GNNs was proposed to tackle irregular data structures, such as graphs and 3D point clouds, and have been applied from social networks to biomedicine in recent years \cite{tomas2017semi,william2017inductive,peter2018graph,johannes2019predict,felix2019simplifying,chen2020simple}. Since large-scale datasets and comprehensive architectures are required in specific applications, how to design lightweight GNNs becomes an important and challenging research problem. To the best of our knowledge, Yang et al. \cite{yang2020distilling} made the first attempt to adopt the idea of knowledge distillation to compress GNNs. They proposed a loss called local structure preserving (LSP) to make the student model imitate the teacher model in pairwise similarity of node features. 
After that, several subsequent variants extended the local structure into the global structure or redesigned an indicator to capture more complex graph structure information \cite{joshi2021representation,yang2021extract,deng2021graphfree}.

Almost all current methods about knowledge extraction on GNNs are offline (see Fig. \ref{fig:overview}(a)), i.e., transferring knowledge from a pre-trained teacher model to a student model. In some complex tasks, however, a pre-trained teacher model is not always available for various reasons, such as training cost, privacy, etc. 
Inspired by recent works that the student performance could be boosted by learning from other peers \cite{zhang2018deep,anil2018large,chen2020online}, as shown in Fig. \ref{fig:overview}(b), we propose a novel online knowledge distillation framework specially designed for GNNs, where one student model learns the structure information extracted from another simultaneously trained counterpart in an alternating training procedure. 

Furthermore, although different model layers represent different semantic spaces, the same node in each layer should have similar local structures \cite{Pei2020GeomGCNGG}. 
The previous proposed LSP module \cite{yang2020distilling} only manually selects one layer output of the teacher and student models to be aligned, which inevitably ignores the information contained in other layers. Therefore, we propose a strategy called \textit{Alignahead}, where each layer of one student model learns from the \textit{next layer} of another student model. In this way, the structure information will be spread over all layers of the two student models after several alternating iterations and each layer can capture the local structure information in other layers. We will give a detailed analysis about this phenomenon on section \ref{subsec:analysis}. To verify the effectiveness of Alignahead, we conduct experiments on different public datasets and different GNN architectures. The main contributions can be summarized as follows:
\begin{itemize}
\item We design an online knowledge distillation framework specifically for GNNs, where two student models alternately learn the structure information extracted from each other without a pre-trained teacher model in advance.
\item A novel cross-layer knowledge distillation strategy called Alignahead is proposed. Each student layer is aligned ahead with the layer in different depth of another student model, which results in the structure information spreading over all layers after several training iterations.
\item Experimental results on various public datasets and model architectures demonstrate that our proposed collaborative training framework is effective to improve the performance and convergence speed of student models.
\end{itemize}

\section{Related Work}
\subsection{Knowledge Distillation}
Knowledge distillation was first proposed by \cite{hinton2015distilling}, which used soft targets to transfer knowledge from the teacher model to the student model. The soft targets, i.e., the teacher model predictions can capture the relationship among classes. In addition to the output layer, there is also a lot of works to extract knowledge from the intermediate layers \cite{romero2015fitnets,zagoruyko2017paying,chen2021cross,zhang2021confidence}. FitNet \cite{romero2015fitnets} added an additional fully-connected layer to the intermediate student output and forced it to match the intermediate feature maps of the teacher model. Attention transfer \cite{zagoruyko2017paying} extracted attention maps instead of features. SemCKD \cite{chen2021cross} used an attention mechanism to automatically assign the appropriate target layer of the teacher model to each student layer. 

However, a powerful teacher is not always available. The recently proposed online knowledge distillation discards pre-trained teacher model, and trains multiple student models by aggregating and aligning their outputs \cite{anil2018large,zhang2018deep,chen2020online}. Deep mutual learning \cite{zhang2018deep} trained a set of student models collaboratively by learning from each other throughout the training process, which also increases the model robustness. OKDDip \cite{chen2020online} performed two-level  distillation by introducing multiple auxiliary peers and one team leader, which achieved good performance without increasing computational complexity in inference. 

\subsection{Graph Neural Networks.}
Recently, graph neural networks have achieved promising results in processing graph data \cite{tomas2017semi,william2017inductive,peter2018graph,johannes2019predict,felix2019simplifying,chen2020simple}. Kipf et al. \cite{tomas2017semi} proposed the Graph Convolution Network (GCN), which propagates node features by spectral graph convolutions. GraphSAGE \cite{william2017inductive} 
further improved the scalability by sampling neighbor nodes and aggregating their features. Graph Attention Network (GAT) \cite{peter2018graph} adopted the attention mechanism to aggregate local features of neighbor nodes with the automatically learned weights. Different from the encouraging progress made in GNNs architecture design, the purpose of this paper is to design a novel online knowledge distillation framework applicable to various GNNs.

\begin{figure*}[ht]
\centering
\includegraphics[width=\linewidth]{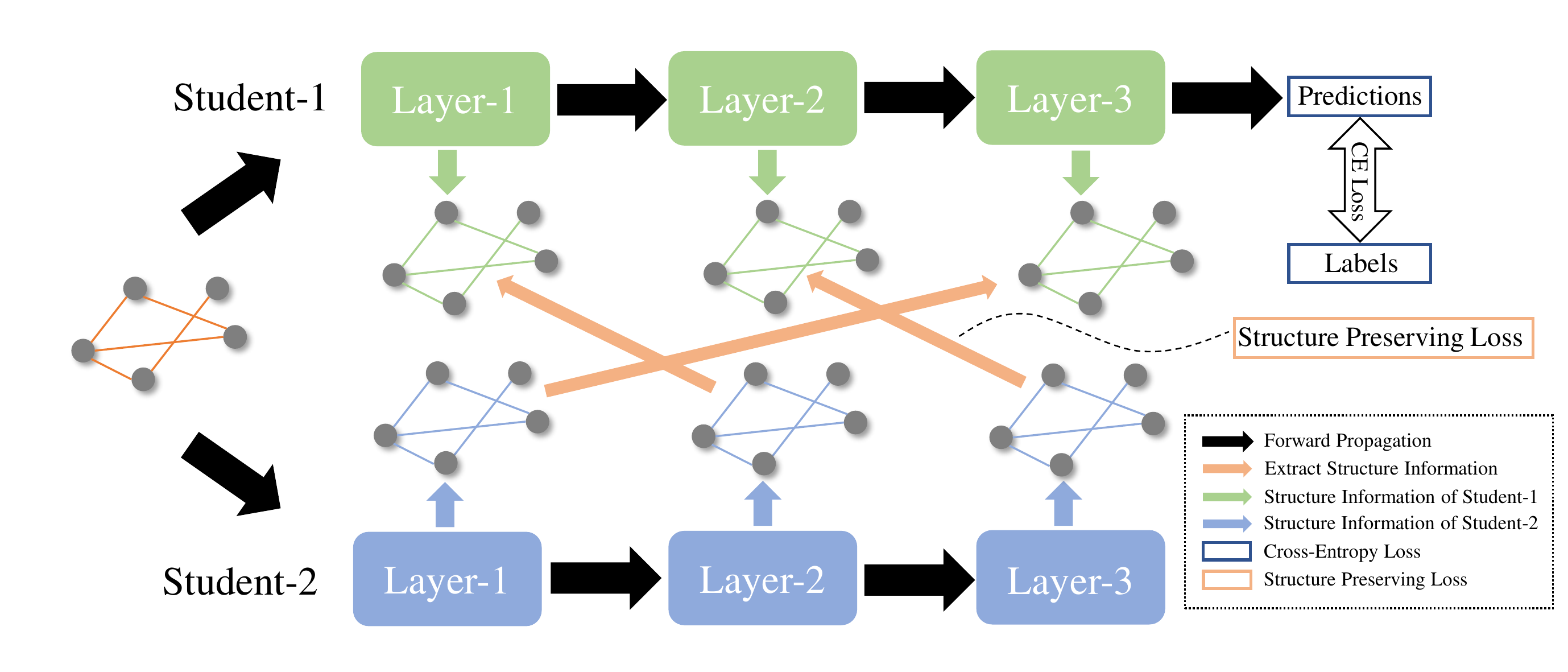}
\caption{Illustration of two alternately trained student models. Our Alignahead strategy makes local structure information to be spread among different layers. The loss function consists of two parts: the cross entropy loss between model predictions and labels, and the cross-layer local structure preserving loss.}
\label{fig:architecture}
\end{figure*}

\subsection{Knowledge Distillation on Graph Neural Networks.}
Yang et al. \cite{yang2020distilling} first tried to extract knowledge from graph neural networks. They designed a module called local structure preserving (LSP) to force the local structure in the intermediate student layer to be consistent with that of teacher model. Joshi et al. \cite{joshi2021representation} extended the LSP module to global structure preservation module and showed that can extract the teacher knowledge better. Yang et al. \cite{yang2021extract} proposed a well-designed student model as a combination of label propagation and feature based multi-layer perceptrons to make full use of the knowledge in a pre-trained teacher model. \cite{deng2021graphfree} considered to extract structural knowledge from GNNs without graph data by using the multivariate Bernoulli distribution to model the graph topology. \cite{Chen2021OnSD} designed a regularization to adaptively transfer knowledge within a single GNN model.

As far as we know, the existing KD methods on GNNs are almost in an offline way. Although \cite{Chen2021OnSD} does not require an extra teacher model, it just transfers knowledge from shallow layers to deep layers. In contrast, we propose a novel online cross-layer knowledge distillation framework to train two student models alternately. With the help of our developed Alignahead strategy, one student model learns the structure information extracted from all layers of itself and other peers after several training iterations to improve its performance.

\section{Method}
This section is organized into three subsections. In section \ref{subsec:lsp}, we give a brief description about the principle of local structure preserving. 
In section \ref{subsec:okd}, we introduce our online knowledge distillation framework with \textit{Alignahead} in detail
and explain why the structure information will flow from layer to layer in section \ref{subsec:analysis}.
\subsection{Local Structure Preserving (LSP)}
\label{subsec:lsp}
LSP is the first knowledge distillation module specifically designed for GNNs, which forces the local structure of the graph data learned by the student model to be similar to that of the teacher model \cite{yang2020distilling}. 
The similarity $L_{i,j}$ between each node $i$ and its neighbor nodes $j|\left(i,j\right)\in \varepsilon$ in the feature space is calculated with one of the following kernel functions and then normalized with a softmax function 
\begin{equation}
D\left(z_{i},z_{j}\right) = 
\begin{cases}
\left\|z_{i}-z_{j} \right\|_{2}^{2} & Euclidean \\ 
z_{i}\cdot z_{j} & Linear \\
\left(z_{i}\cdot z_{j}+c \right)^{d} & Poly\\
e^{-\frac{1}{2\sigma }\left\|z_{i}-z_{j} \right\|^{2}} & RBF
\end{cases} 
\end{equation}
\begin{equation}
L_{i,j} = \frac{e^{D\left(z_{i},z_{j}\right)}}{\sum_{j|\left(i,j\right)\in \varepsilon\, \left(e^{D\left(z_{i},z_{j}\right)}\right)}}\,,
\end{equation}
where $z_{i}$ and $z_{j}$ are the node features. The local structure $L_{i}$ of node $i$ is expressed as the probability distribution of similarity between node $i$ and the nodes in its neighborhood. 

The student model is trained with KL-divergence to mimic the local structure of the teacher model in the feature space:
\begin{equation}
\mathcal{L}_{lsp} = \frac{1}{N}\sum_{i=1}^{N}D_{KL}\left(L_{i}^{T}||L_{i}^{S}\right).
\label{eq:lsp}
\end{equation}

\begin{algorithm}
\caption{Alignahead for collaborative distillation}
\KwIn{The parameters $\theta _{1}$ and $\theta _{2}$ of the student $S_{1}$ and $S_{2}$; The calculated local structure by LSP module;
hyper-parameter $\alpha$ and label $y$}
\KwOut{$S_{1}$ and $S_{2}$ with excellent performance.}
Initialization parameters $\theta _{1}$ and $\theta _{2}$.

\While{$epochs \le max\_epoch$}
{
    // Training loss of $S_{1}$ 
    
    Obtain the structure preserving loss $\mathcal{L}_{str}^{S_{1}}$ with Equ. (\ref{eq:strloss}) and Equ. (\ref{eq:strloss1}).
    
    Obtain the cross-entropy loss   $\mathcal{L}_{ce}^{S_{1}}\left(y, p_{1}\right)$.
         
    Construct the total loss for $S_{1}$ as Equ. (\ref{eq:total-loss}).
    
    // Training loss of $S_{2}$ 
    
    Obtain the structure preserving loss $\mathcal{L}_{str}^{S_{2}}$ with Equ. (\ref{eq:strloss}) and Equ. (\ref{eq:strloss2}).
        
    Obtain the cross-entropy loss:   
    $\mathcal{L}_{ce}^{S_{2}}\left(y, p_{2}\right)$.
         
    Construct the total loss for $S_{2}$ as Equ. (\ref{eq:total-loss}).
    
    // Alternating training of $S_{1}$ and $S_{2}$ 
    
    \textbf{Update} the parameter $\theta _{1}$ while keeping $\theta _{2}$ fixed.
    
    \textbf{Update} the parameter $\theta _{2}$ while keeping $\theta _{1}$ fixed.
        
}
\end{algorithm}

\subsection{Online Knowledge Distillation with Alignahead}
\label{subsec:okd}
\begin{figure*}[ht]
  \centering
  \includegraphics[width=\linewidth]{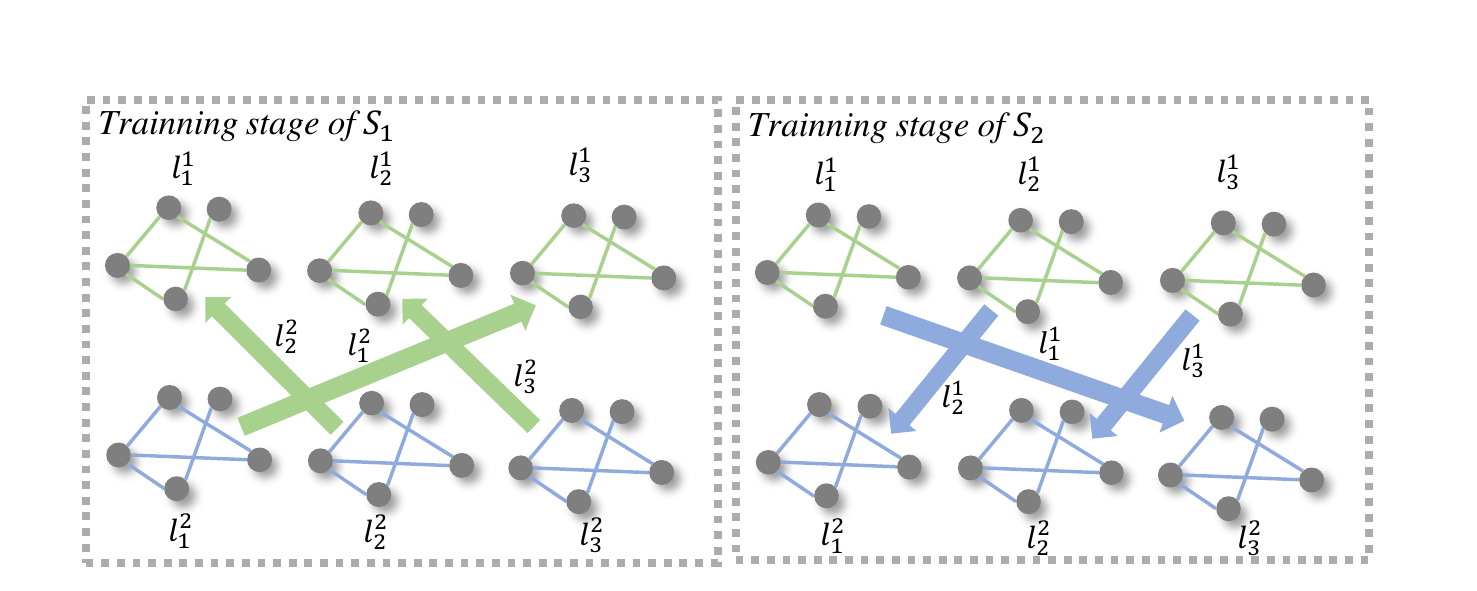}
  \caption{The arrows represent the information flow in alternating training of two student models.}
  \label{fig:flow}
\end{figure*}

\begin{table*}[ht]
\centering
\caption{After six iterations, each student layer captures the structure information from all the other layers. One iteration includes the alternating training of Student-1 and student-2}
\renewcommand{\arraystretch}{1}
\resizebox{0.7\textwidth}{!}{%
\begin{tabular}{lllllllllllllll}
\toprule
\multicolumn{2}{c}{Iteration number} & ~~~0 & \multicolumn{2}{c}{1} & \multicolumn{2}{c}{2} & \multicolumn{2}{c}{3} & \multicolumn{2}{c}{4} & \multicolumn{2}{c}{5} & \multicolumn{2}{c}{6} \\ 
\midrule
\multicolumn{2}{c}{Training stage} & Initial & $S_{1}$ & \multicolumn{1}{c}{$S_{2}$} & $S_{1}$ & \multicolumn{1}{c}{$S_{2}$} & $S_{1}$ & \multicolumn{1}{c}{$S_{2}$} & $S_{1}$ & \multicolumn{1}{c}{$S_{2}$} & $S_{1}$ & \multicolumn{1}{c}{$S_{2}$} & $S_{1}$ & \multicolumn{1}{c}{$S_{2}$} \\
\midrule
Student-1 & Layer-1 & \multicolumn{1}{l|}{\bm{~~$l_{1}^{1}$}}& \bm{$l_{2}^{2}$} & \multicolumn{1}{l|}{$l_{2}^{2}$} &  \bm{$l_{3}^{1}$}& \multicolumn{1}{l|}{$l_{3}^{1}$} &  \bm{$l_{1}^{2}$}& \multicolumn{1}{l|}{$l_{1}^{2}$} &  \bm{$l_{2}^{1}$}& \multicolumn{1}{l|}{$l_{2}^{1}$} &  \bm{$l_{3}^{2}$}& \multicolumn{1}{l|}{$l_{3}^{2}$} &  \bm{$l_{1}^{1}$}& $l_{1}^{1}$ \\ \cmidrule(l){2-15}
& Layer-2 & \multicolumn{1}{l|}{~~$l_{2}^{1}$} & $l_{3}^{2}$ & \multicolumn{1}{l|}{$l_{3}^{2}$} & $l_{1}^{1}$ & \multicolumn{1}{l|}{$l_{1}^{1}$} &  $l_{2}^{2}$& \multicolumn{1}{l|}{$l_{2}^{2}$} &  $l_{3}^{1}$& \multicolumn{1}{l|}{$l_{3}^{1}$} &  $l_{1}^{2}$& \multicolumn{1}{l|}{$l_{1}^{2}$} &  $l_{2}^{1}$& $l_{2}^{1}$ \\ \cmidrule(l){2-15}
& Layer-3 & \multicolumn{1}{l|}{~~$l_{3}^{1}$} & $l_{1}^{2}$ & \multicolumn{1}{l|}{$l_{1}^{2}$} &  $l_{2}^{1}$& \multicolumn{1}{l|}{$l_{2}^{1}$} &  $l_{3}^{2}$& \multicolumn{1}{l|}{$l_{3}^{2}$} &  $l_{1}^{1}$& \multicolumn{1}{l|}{$l_{1}^{1}$} &  $l_{2}^{2}$& \multicolumn{1}{l|}{$l_{2}^{2}$} &  $l_{3}^{1}$& $l_{3}^{1}$ \\ \midrule
Student-2 & Layer-1 & \multicolumn{1}{l|}{~~$l_{1}^{2}$} & $l_{1}^{2}$ & \multicolumn{1}{l|}{$l_{2}^{1}$} &  $l_{2}^{1}$& \multicolumn{1}{l|}{$l_{3}^{2}$} &  $l_{3}^{2}$& \multicolumn{1}{l|}{$l_{1}^{1}$} &  $l_{1}^{1}$& \multicolumn{1}{l|}{$l_{2}^{2}$} &  $l_{2}^{2}$& \multicolumn{1}{l|}{$l_{3}^{1}$} & $l_{3}^{1}$ & $l_{1}^{2}$ \\ \cmidrule(l){2-15} 
 & Layer-2 & \multicolumn{1}{l|}{~~$l_{2}^{2}$} & $l_{2}^{2}$ & \multicolumn{1}{l|}{$l_{3}^{1}$} &  $l_{3}^{1}$& \multicolumn{1}{l|}{$l_{1}^{2}$} &  $l_{1}^{2}$& \multicolumn{1}{l|}{$l_{2}^{1}$} &  $l_{2}^{1}$& \multicolumn{1}{l|}{$l_{3}^{2}$} &  $l_{3}^{2}$& \multicolumn{1}{l|}{$l_{1}^{1}$} & $l_{1}^{1}$ & $l_{2}^{2}$ \\ \cmidrule(l){2-15} 
 & Layer-3 &  \multicolumn{1}{l|}{~~$l_{3}^{2}$} & $l_{3}^{2}$ & \multicolumn{1}{l|}{$l_{1}^{1}$} &  $l_{1}^{1}$& \multicolumn{1}{l|}{$l_{2}^{2}$}&  $l_{2}^{2}$&  \multicolumn{1}{l|}{$l_{3}^{1}$}& $l_{3}^{1}$&  \multicolumn{1}{l|}{$l_{1}^{2}$}&  $l_{1}^{2}$&  \multicolumn{1}{l|}{$l_{2}^{1}$}& $l_{2}^{1}$ & $l_{3}^{2}$ \\ 
\bottomrule
\end{tabular}%
}
\label{tab: alignahead}
\end{table*}

The LSP module does successfully transfer the local structure information of the teacher model to the student model. However, due to several reasons like training cost and privacy, the pre-trained teacher model is not always available. 

We design an online knowledge distillation framework specially for GNNs, which does not require the assistance of a pre-trained teacher model, but only requires several student models to learn from each other, resulting in a better performance than training with ground-truth labels alone. 
In addition, we develop a cross-layer knowledge distillation strategy called \textit{Alignahead}, where one student layer is aligned ahead with the layer in different depth of the another student model. In each round of alternate training, the structure information of each student layer is transferred to the previous layer of another student model, and finally the structure information will spread over all layers. Our collaborative training framework with two student models are illustrated in Fig. \ref{fig:architecture} and discussed in the following paragraphs.

Suppose we have two student models $S_{1}$ and $S_{2}$ with the same architecture, the intermediate layer number is $H$, we use $l_{i,j}^{S_{1}}$ and $l_{i,j}^{S_{2}}$ to represent the local structure of node $j$ in the $i$-th layer. For $S_{1}$, its local  structure of $i$-th layer is required to match the $\left(i+1\right)$-th layer of $S_{2}$ and the final layer $H$ is required to match the first layer. The structure preserving loss is calculated as the sum of KL divergence loss over all layers
\begin{equation}
\mathcal{L}_{str}^{S_{1}} = \sum_{i=1}^{H}L^{S_{1}}_{{l}_{i}},\quad
\mathcal{L}_{str}^{S_{2}} = \sum_{i=1}^{H}L^{S_{2}}_{{l}_{i}},
\label{eq:strloss}
\end{equation}
where 
\begin{equation}
\mathcal{L}^{S_{1}}_{{l}_{i}} = \frac{1}{N}\sum_{j=1}^{N}D_{KL}\left(l_{i+1,j}^{S_{2}}||l_{i,j}^{S_{1}}\right),\label{eq:strloss1}
\end{equation}
\begin{equation}
\mathcal{L}^{S_{2}}_{{l}_{i}} = \frac{1}{N}\sum_{j=1}^{N}D_{KL}\left(l_{i+1,j}^{S_{1}}||l_{i,j}^{S_{2}}\right).\label{eq:strloss2}
\end{equation}

The total loss of $S_{1}$ and $S_{2}$ is formulated as:
\begin{align}
\mathcal{L}_{1} = \mathcal{L}_{ce}^{S_{1}}\left(y, p_{1}\right)+\alpha L_{str}^{S_{1}}, \quad
\mathcal{L}_{2} = \mathcal{L}_{ce}^{S_{2}}\left(y, p_{2}\right)+\alpha L_{str}^{S_{2}},\label{eq:total-loss}
\end{align}
where $\mathcal{L}_{ce}^{S_{1}/S_{2}}$ is a cross entropy loss function calculated with the student prediction $p_{1}/p_{2}$ and the label $y$. Hyper-parameter $\alpha$ is used to balance two parts of losses. Two student models $S_{1}$ and $S_{2}$ are trained alternately as shown in Algorithm 1.

\subsection{Why structure information spread over all layers?}
\label{subsec:analysis}
In this section, we explain why the structure information spreads over all layers. Fig. \ref{fig:flow} reveals the flow of structure information during an alternating training procedure. 
It can be seen that each student layer is aligned ahead with the next layer of another student model and the last layer is to match the first layer specially. 
In this way, the two student models exchange structure information with each other and propagate it in different layer depth. 

Table \ref{tab: alignahead} shows the distribution of structure information of two student models with three hidden layers in six iterations. 
Theoretically, the structure information of each layer is circulated once inside the two student models. 
Take the layer-1 of the Student-1 model as an example, it gathers the structure information in $l_{2}^{2}$, $l_{3}^{1}$, $l_{1}^{2}$, $l_{2}^{1}$, $l_{3}^{2}$ and $l_{1}^{1}$ one-by-one, and returns to the initial state in the sixth iteration. That is to say, the layer-1 of the Student-1 model indeed collects the structure information from all layers of two student models within the first six iterations. Similar phenomenon also happens in other layers, making the structure information spread over all layers.

\begin{table*}[ht]
\centering
\renewcommand{\arraystretch}{1}
\caption{Summary of the datasets. The PPI dataset is used for inductive learning, and the others are used for transductive learning.}
\resizebox{0.9\linewidth}{!}{
\begin{tabular}{cccccccc}
\toprule
\textbf{Dataset} & \textbf{\#Graphs} & \textbf{\#Nodes} & \textbf{\#Edges} & \textbf{\#Features} & \textbf{\#Classes} & \textbf{\#Task} &\textbf{\#Metric} \\ 
\midrule
PPI             & 24       & 56944   & 818716  & 50         & 121 (multilabel) & Inductive &F1 score \\
Coauthor-CS     & 1        & 18333   & 81894   & 6805       & 15             & Transductive & Accuracy \\
Coauthor-Physics    & 1        & 34493   & 247962  & 8415       & 5       & Transductive        & Accuracy \\
Amazon-Computer & 1        & 13381   & 245778  & 767        & 10          & Transductive    & Accuracy \\
Amazon-Photo    & 1        & 7487    & 119043  & 745        & 8           & Transductive    & Accuracy \\ 
\bottomrule
\end{tabular}
}
\label{tab:dataset}
\end{table*}

\section{Experiment}
Here, we first introduce five benchmark datasets, the experimental settings and three popular graph neural network architectures adopted in our experiments in section \ref{subsec:datasets-models}. Section \ref{compared} lists the compared methods. Section \ref{subsec:inductive} and \ref{transductive} detail the experimental results with different methods in two different tasks.
Finally, we explore the effect of student numbers and hyper-parameter choices on the model performance.

\subsection{Datesets and student models}
\label{subsec:datasets-models}
Five benchmark datasets are used in our experiments, and their summary are shown in Table \ref{tab:dataset} and listed as follows:

\begin{itemize}
\item PPI \cite{marinka2017predicting} consists of 24 graphs corresponding to different human tissues, where 20 graphs are used for training, two graphs are used for validation, and another 2 graphs are used for testing. The average node numbers for each graph is 2372. Each node has 50 features, consisting of positional gene sets, motif gene sets, and immunological signatures. The number of labels is 121 and a node can have multiple labels at the same time. PPI dataset is used for inductive learning.
\item Coauthor-CS and Coauthor-Physics \cite{shchur2018pitfalls} are co-author graphs which belong to the field of computer science and physics, respectively. Authors are denoted as nodes, and edges indicate whether there is a cooperative relationship between two authors. Node features represent the keywords of each author's paper, and labels indicate each author's most active research field. Coauthor-CS/Physics datasets are used for transductive learning.
\item Amazon-Computer and Amazon-Photo \cite{shchur2018pitfalls,julian2015image} are parts of Amazon co-purchase graph, where the nodes represent products, and the edges represent whether two products are always purchased together. Node features represent product reviews encoded in bag-of-words, and labels are given by product categories. These two datasets are used for transductive learning.
\end{itemize}

The adopted models are briefly listed as follows:
\begin{itemize}
\item GCN \cite{tomas2017semi} propagates node features by defining spectral graph convolutions on graph data. In the experiments, we adopt a 3-layer GCN model.
\item SAGE \cite{william2017inductive} updates node features by sampling neighbor nodes and aggregating their information. SAGE-GCN, SAGE-mean and SAGE-pool are used as student models.
\item GAT \cite{peter2018graph} introduces the attention mechanism to automatically assign weights to nodes' neighbors. In the PPI dataset, we adopt GAT with different layers and dimensions for the teacher model and the student model.
\end{itemize}

In the case of the PPI dataset, we train the GAT \cite{peter2018graph} student model on the visible graph to make predictions on completely invisible graphs. We refer to the experiment settings of \cite{yang2020distilling}, where the learning rate, weight decay, optimizer and epochs are 0.005, 0, Adam and 300, respectively. We adopt the RBF kernel function for distance calculation in LSP module, and $\sigma$ is 100. 
While on other four datasets, we apply GCN \cite{tomas2017semi} and three variants of SAGE \cite{william2017inductive} as student models to predict the unknown nodes on the visible graph. 
We adopt one set of experiment settings from \cite{yang2021extract}, where the learning rate, weight decay, optimizer and epochs are 0.001, 0.0005, Adam and 200, respectively, and the Euclidean kernel function is adopted for all models. All our experiments are performed on an NVIDIA 2080Ti GPU. 

\subsection{Compared methods}
\label{compared}
The results of three methods are presented in comparison:
\begin{itemize}
\item \textit{Self} means the student model is trained with labels alone.
\item \textit{OC} is the ablation model of \textit{Alignahead}. It keeps the setting of online knowledge distillation but changes cross-layer matching to one-to-one correspondence, which means each layer of the student model is matched with the corresponding layer of the peer model. We report the higher metric of the two student models.
\item \textit{Alignahead} means the student models are trained with our proposed online cross-layer distillation framework, as shown in Algorithm 1.
\end{itemize}
We report the best student model's F1 score in Section \ref{subsec:num} and \ref{subsec:sen}, named as max F1 score. 
 
\subsection{Inductive learning}
\label{subsec:inductive}
On the PPI dataset, our goal is to use the visible graphs to train the GAT student model to predict the invisible graphs and achieve node classification. Similar to the previous work \cite{yang2020distilling}, we adopt a teacher model that is shallower but has more hidden features than the student model, as shown in Table \ref{tab:SandT}. 

\begin{table}[ht]
\centering
\renewcommand{\arraystretch}{1.2}
\caption{The model structures used on PPI dataset.}
\resizebox{0.95\linewidth}{!}{
\begin{tabular}{ccccc}
\toprule
\textbf{Model}   & \textbf{Layers} & \textbf{Attention} \textbf{heads} & \textbf{Feature maps} & \textbf{Params}\\
\midrule
Teacher & 4      & 6, 6, 6, 6           & 256, 256, 256     & 11.75M\\
Student & 5      & 3, 3, 3, 3, 3         & 64, 64, 64, 64 & 0.37M\\
\bottomrule
\end{tabular}
}
\label{tab:SandT}
\end{table}
\begin{table}[ht]
\caption{Experimental results on the PPI dataset.}
\renewcommand{\arraystretch}{1.1}
\centering
\resizebox{0.8\linewidth}{!}{
\begin{tabular}{cc}
\toprule
\textbf{Model}                        & \textbf{F1 score}        \\
\midrule
Teacher\_self                & 0.9871          \\
Student\_self                & 0.9736          \\
Student\_LSP \cite{yang2020distilling}                & 0.9751         \\
Student\_OC   & 0.9756 \\
(Student1, Student2)\_Alignahead & (\textbf{0.9766}, 0.9753)\\
\bottomrule
\end{tabular}
}
\label{tab:PPIresult}
\end{table}

\begin{table*}[ht]
\centering
\caption{Experimental results on the four datasets. Here we use two student models with the same architecture.}
\resizebox{\textwidth}{!}{%
\begin{tabular}{ccccc|cccc|cccc|cccc}
\toprule
\multicolumn{1}{c|}{Datasets} & \multicolumn{4}{c|}{Coauthor-CS} & \multicolumn{4}{c|}{Coauthor-Physics} & \multicolumn{4}{c|}{Amazon-Computer} & \multicolumn{4}{c}{Amazon-Photo} \\ \midrule 
\multicolumn{1}{c|}{Methods} & \multicolumn{1}{c|}{self} & \multicolumn{1}{c|}{OC} & \multicolumn{2}{c|}{Alignahead} & \multicolumn{1}{c|}{self} & \multicolumn{1}{c|}{OC} & \multicolumn{2}{c|}{Alignahead} & \multicolumn{1}{c|}{self} & \multicolumn{1}{c|}{OC} & \multicolumn{2}{c|}{Alignahead} & \multicolumn{1}{c|}{self} & \multicolumn{1}{c|}{OC} & \multicolumn{2}{c}{Alignahead} \\ \midrule
\multicolumn{1}{c|}{GCN} & \multicolumn{1}{c|}{0.9137} & \multicolumn{1}{c|}{0.9144} & \textbf{0.9152} & 0.9152 & \multicolumn{1}{c|}{0.9638} & \multicolumn{1}{c|}{0.9638} & \textbf{0.9645} & 0.9645 & \multicolumn{1}{c|}{0.8697} & \multicolumn{1}{c|}{0.8580} & \textbf{0.8747} & 0.8747 & \multicolumn{1}{c|}{0.9117} & \multicolumn{1}{c|}{0.9138} & \textbf{0.9158} & 0.9158 \\
\multicolumn{1}{c|}{SAGE-mean} & \multicolumn{1}{c|}{0.9347} & \multicolumn{1}{c|}{0.9354} & 0.9370 & \textbf{0.9377} & \multicolumn{1}{c|}{0.9705} & \multicolumn{1}{c|}{0.9711} & 0.9712 & \textbf{0.9719} & \multicolumn{1}{c|}{0.8986} & \multicolumn{1}{c|}{0.9008} & 0.9022 & \textbf{0.9051} & \multicolumn{1}{c|}{0.9507} & \multicolumn{1}{c|}{0.9507} & \textbf{0.9528} & 0.9528 \\
\multicolumn{1}{c|}{SAGE-GCN} & \multicolumn{1}{c|}{0.9197} & \multicolumn{1}{c|}{0.9205} & \textbf{0.9220} & 0.9212 & \multicolumn{1}{c|}{0.9625} & \multicolumn{1}{c|}{0.9625} & \textbf{0.9645} & 0.9638 & \multicolumn{1}{c|}{0.8595} & \multicolumn{1}{c|}{0.8624} & 0.8639 & \textbf{0.8675} & \multicolumn{1}{c|}{\textbf{0.9302}} & \multicolumn{1}{c|}{0.9280} & 0.9240 & 0.9179 \\
\multicolumn{1}{c|}{SAGE-pool} & \multicolumn{1}{c|}{0.8950} & \multicolumn{1}{c|}{0.8927} & \textbf{0.8980} & 0.8957 & \multicolumn{1}{c|}{0.9531} & \multicolumn{1}{c|}{0.9551} & \textbf{0.9571} & 0.9558 & \multicolumn{1}{c|}{0.9005} & \multicolumn{1}{c|}{0.9044} & \textbf{0.9051} & 0.9038 & \multicolumn{1}{c|}{0.9546} & \multicolumn{1}{c|}{0.9487} & 0.9548 & \textbf{0.9568} \\ 
\bottomrule
\end{tabular}%
}
\label{tab:trans-result}
\end{table*}

The experimental results are shown in Table \ref{tab:PPIresult}. It can be seen that student models trained with our proposed Alignahead consistently achieve better performance compared to those trained with labels or only the LSP module. The comparison with the results of one-to-one correspondence (OC) demonstrates Alignahead can indeed help the student layer learn global semantics by flowing structure information.

\begin{figure}[ht]
  \centering
  \includegraphics[width=0.9\linewidth]{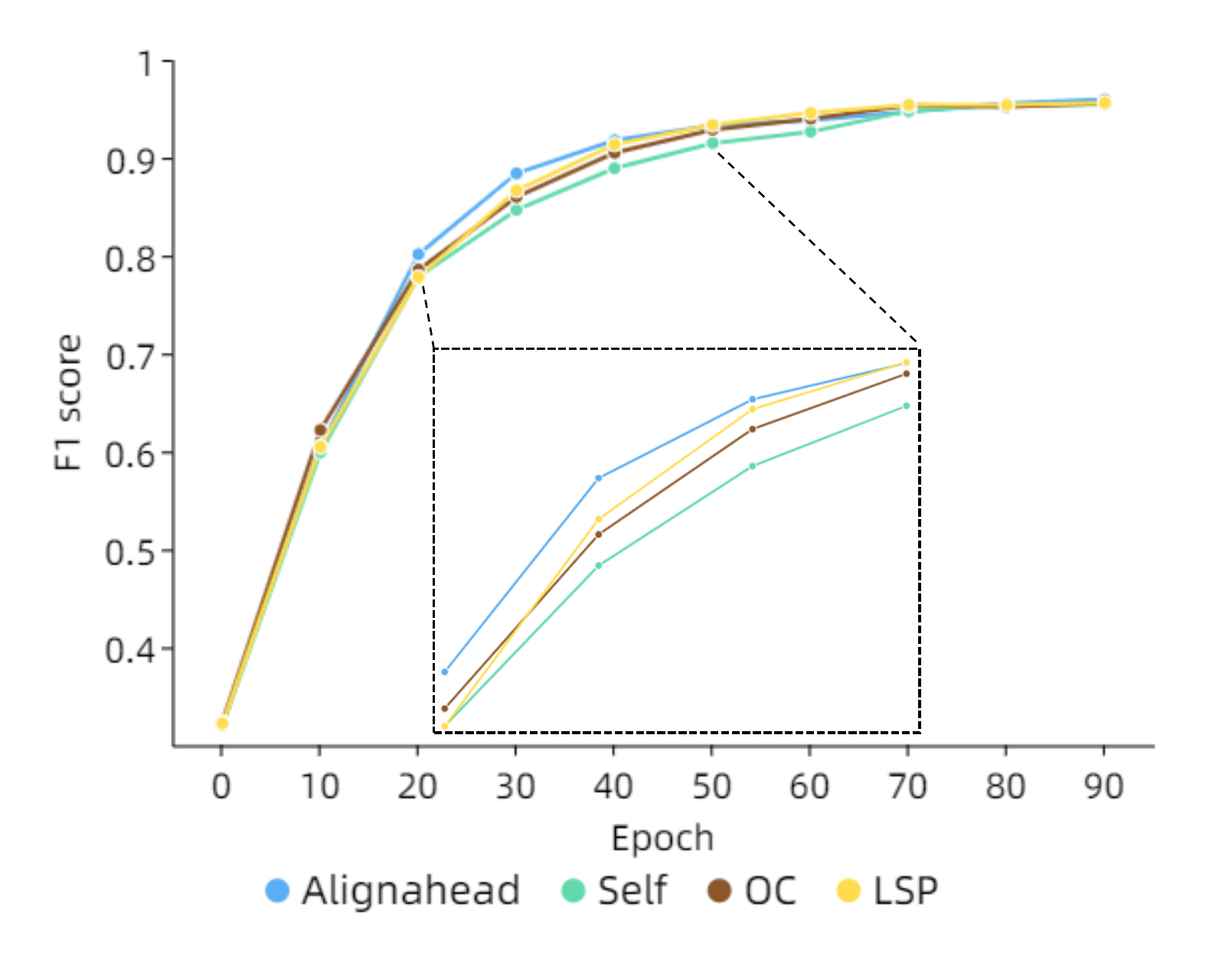}
  \caption{F1 score of different methods during training.}
  \label{fig:convergence}
\end{figure}

Next, we explore the convergence speed of student models training with different methods, and the results are shown in Fig. \ref{fig:convergence}. 
%We intercept the first 100 epochs and draw a line graph with every 10 epochs as a marker point. 
As we can see, Alignahead is almost always in the lead at the first 100 epochs and the performances of LSP and OC are similar, which indicates the effectiveness of our cross-layer knowledge distillation strategy in speeding up the model convergence. Compared with LSP, though without the supervision of a pre-trained teacher model, 
the student model trained with OC strategy can extract the structure information from each layer of the peers rather than a certain layer, resulting the convergence speed acceleration. Even so, using only one layer of structure information is not enough. 
Our strategy \textit{Alignahead} spreads the structure information over all layers in the training process to make each student layer get richer and more comprehensive information, which makes our model converges faster compared with OC strategy.

\subsection{Transductive learning}
\label{transductive}
We conduct experiments with GCN as well as three variants of SAGE (SAGE-mean, SAGE-GCN and SAGE-pool) on the Coauthor-CS, Coauthor-Physics, Amazon-Computer and Amazon-Photo datasets. The goal is to train these student models to classify unknown nodes on the visible graph. For GCN and three variants of SAGE, we adopt a three-layer framework with 128 dimensional features in each layer. Two student models with the same architecture are used for training, and the purpose is to verify whether our framework can help them get better performance than training alone with the ground-truth labels. 

As we can see in Table \ref{tab:trans-result}, almost all student models have improved performance with our Alignahead strategy. Among them, SAGE-GCN has the largest improvement on Amazon-Computer, and the accuracy has increased by 0.8$\%$, but there is no improvement on Amazon-Photo. Overall, the three variants of SAGE outperform the GCN model on the four datasets. On the Amazon-Computer dataset, the four student models have the largest improvement, reaching an average of 0.6$\%$. 
In addition, the one-to-one correspondence (OC) strategy does not seem to work in some experiments and their structure preserving loss are almost unchanged, which means that this simple strategy may cause the training quickly fall into a local optimum.
While our proposed Alignahead can capture the structure information of all layers for each student layer to achieve better results.

\subsection{The number of student models}
\label{subsec:num}
In this section, we explore the effect of the number of student models on the experimental results. If the student number is greater than 2, each model will capture local structure information from all remaining models. The $i$-th layer's structure preserving loss of the $k$-th model becomes
\begin{align}
L_{l_{i}}^{S_{k}} = \frac{1}{M-1}\frac{1}{N}\sum_{p=1 \vee p\neq k}^{M}\sum_{j=1}^{N}D_{KL}\left(l_{i+1,j}^{S_{p}}||l_{i,j}^{S_{k}}\right),
\end{align}
where $M$ is the number of student models.
To make the student models have more room for improvement, we use the GAT with a smaller architecture on PPI dataset. The student model is a four-layer GAT, and the number of attention heads and feature dimensions of each layer are [2,2,2,2] and [42,42,42]. We use the same parameter settings as in Section \ref{subsec:inductive}. 

As shown in Table \ref{tab:model-number}, increasing the student numbers does improve the model performance especially when it is less than 4. However, when the number increases further, the Max F1 score hardly changes, which indicates that the model performance has reached its limit. 
%Thus, we think it doesn't make sense to continue to increase the number of student models. 
Since the model training time linearly increases as the student numbers increase, two or three student models are generally appropriate in practice.

\begin{table}[ht]
\centering
\renewcommand{\arraystretch}{1.2}
\caption{The effects of different student numbers on the PPI dataset. All student models are GAT with the same architecture. Metric is the max F1 score.}
\resizebox{0.95\linewidth}{!}{%
\begin{tabular}{cccccc}
\toprule
\multicolumn{1}{c}{Number of models} & 2 & 3 & 4 & 5 & 6 \\ 
\midrule
\multicolumn{1}{c|}{Alignahead}  & 0.8502 & 0.8511 & 0.8523 & 0.8526 & 0.8525 \\
\multicolumn{1}{c|}{OC}  & 0.8482 & 0.8505 & 0.8498 & 0.8504 & 0.8507 \\ 
\bottomrule
\end{tabular}
}
\label{tab:model-number}
\end{table}

\subsection{Sensitivity analysis}
\label{subsec:sen}
\begin{figure}[ht]
  \centering
  \includegraphics[width=0.9\linewidth]{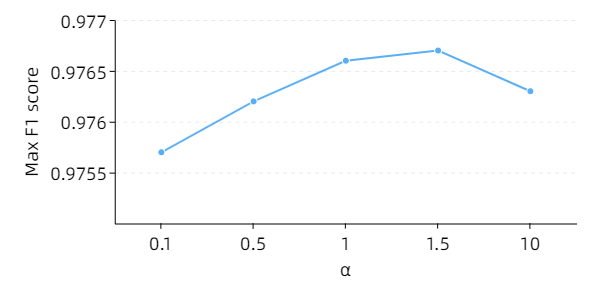}
  \caption{Sensitivity analysis of hyper-parameter $\alpha$ on the PPI dataset. }
  \label{fig:sen}
\end{figure}

In this section, we explore the impact of different values of $\alpha$ on the results. The same model architecture and experiment settings as the Section \ref{subsec:inductive} are used. We take 0.1, 0.5, 1, 1.5 and 10 for $\alpha$, respectively, and report the higher F1 score of the two student models. The results are shown in Fig. \ref{fig:sen}. These experimental results demonstrate the robustness of our model since the all the F1 scores are between 0.975 and 0.977.

\section{Conclusion}
In this paper, we propose an online knowledge distillation framework specifically for graph neural networks, where two student models extract the structure information from each other in an alternating training procedure. We further design the Alignahead strategy to align one student layer with the next layer of another student model, resulting in the structure information spreading over all layers after several iterations. We conduct experiments with GAT, GCN and three variants of SAGE on five datasets: PPI, Coauthor-CS, Coauthor-Physics, Amazon-Computer and Amazon-Photo. The results show that the collaboratively trained student models with the Alignahead technique achieve better performance than the student model training with the ground-truth labels alone. Besides, we find that our framework can accelerate the model convergence speed and increasing the student numbers generally improves the performance further. Our framework also exhibits high robustness with different hyper-parameter settings.

\section{Acknowledgment}
This work is supported by the Starry Night Science Fund of Zhejiang University Shanghai Institute for Advanced Study (Grant No: SN-ZJU-SIAS-001), National Natural Science Foundation of China (Grant No: U1866602). The authors would like to thank anonymous reviewers for their helpful comments.
\bibliographystyle{IEEEtran}
\bibliography{egbib}
\balance
\end{document}